\documentclass{article}
\usepackage{spconf,amsmath,graphicx,hyperref,amssymb,mathtools, tikz-cd, quiver, subcaption}

\title{Scale-Covariant Spiking Wavelets}
\name{Jens Egholm Pedersen$^1$, Tony Lindeberg$^2$, Peter Gerstoft$^1$\thanks{Support from the Novo Nordisk Foundation  (NNF24OC0089302) and the Swedish Research Council (2022-02969) is gratefully acknowledged.}
}
\address{$1\colon$ DTU Electro, Technical University of Denmark, Denmark \\ $2\colon$ Department of Computational Science and Technology, KTH Royal Institute of Technology, Sweden \\*[-2ex]}
\begin{document}
\ninept
\maketitle
\begin{abstract}
% The abstract should appear at the top of the left-hand column of text, about
% 0.5 inch (12 mm) below the title area and no more than 3.125 inches (80 mm) in
% length.  Leave a 0.5 inch (12 mm) space between the end of the abstract and the
% beginning of the main text.  The abstract should contain about 100 to 150
% words, and should be identical to the abstract text submitted electronically
% along with the paper cover sheet.  All manuscripts must be in English, printed
% in black ink.
We establish a theoretical connection between wavelet transforms and spiking neural networks through scale-space theory.
We rely on the scale-covariant guarantees in the leaky integrate-and-fire neurons to implement discrete mother wavelets that approximate continuous wavelets.
A reconstruction experiment demonstrates the feasibility of the approach and warrants further analysis to mitigate current approximation errors.
Our work suggests a novel spiking signal representation that could enable more energy-efficient signal processing algorithms.
\end{abstract}
\section{Introduction}
\label{sec:intro}

The deep learning paradigm is facing challenging scaling issues, currently due to energy consumption and transistor scaling \cite{Mead_2023}.
Biologically-inspired spiking neural networks have been hypothesized to address some limitations in deep learning, including time, space, and energy \cite{Mead_2023, Aimone_2025}.
However, current computational models have not convincingly realized the promising gains, and we lack theories to guide better implementations \cite{pedersen2025covariant, Aimone_2025}.

A key challenge for spiking neural systems is the choice of coding.
Neural systems exhibit singular, digital activations in massively parallel networks that are highly sparse in both space and time \cite{Mead_2023}.
This leaves a vast search space for possible encodings, and discretization is challenging for training spiking networks at scale \cite{pedersen2025covariant}.

Signals are efficiently represented by the wavelet transform \cite{daubechies2006Ten}, based on orthogonal or semi-overlapping bases for signals at varying resolutions and scales.
However, conventional wavelet implementations on digital hardware have fundamental energy constraints in always-on, edge, or real-time systems.
Neuromorphic hardware offers orders-of-magnitude energy improvements through event-driven computation \cite{dampfhoffer2022snns}, but lacks theoretical foundations that connect the neural primitives to signal processing. Establishing that spiking neural networks can implement wavelets enables rigorous, energy-efficient signal processing on neuromorphic substrates.
This work bridges signal processing and neuromorphic computing by leveraging the connection between scale-space representations and wavelets \cite{lindeberg2023time}.
We demonstrate that leaky integrate-and-fire neurons can form mother wavelets, enabling applications in edge sensing (continuous processing with microJoule budgets), audio processing (native spike-domain computation for event-based microphones), and signal compression (exploiting spatio-temporal sparsity). Through standardized interfaces such as the Neuromorphic Intermediate Representation \cite{pedersen2024neuromorphic}, these theoretical foundations translate into deployable primitives on commercial neuromorphic hardware.

The admissibility criteria for a mother wavelet were extended to scale-space representations, a covariant re-parameterization of a signal across varying scales \cite{lindeberg2023time}.
Covariance for signal transformations is an interesting property leveraged in many domains \cite{porcu30YearsSpace2021}, including computer vision \cite{cohenGroupEquivariantConvolutional2016}, PDE solving \cite{smetsPDEBasedGroupEquivariant2023}, graph convolutional neural networks \cite{bekkersb}, and recently for spiking neural networks \cite{pedersen2025covariant}.

In this paper, we leverage the connection between scale-space and wavelets to study spiking wavelets as the basis for mathematically sound and compact representations.

%\subsection{Relation to prior work}
\label{sec:prior}

The relation between wavelets and scale-spaces was formally introduced in \cite{lindeberg2023time} and extended to Gabor kernels in \cite{lindeberg2024time}.
Wavelets have been used in neural networks to model complex processes \cite{alexandridis2013wavelet}.
Spiking networks implement wavelets in experiments described in \cite{zhang2013fast}, and \cite{haghighatshoar2025lowpower} leveraged Hilbert transforms for beamforming using spiking neurons.
Spiking neural networks are brain-inspired (neuromorphic) neural networks that co-locate memory and computation, avoiding the von Neumann bottleneck, and have high performance and low energy consumption. Due to their compact hardware implementations, neuromorphic computing can run machine learning on the edge rather than on a central computer.  Neuromorphic computing is based on impulses and is more energy-efficient \cite{dampfhoffer2022snns}, but challenging to represent on conventional digital systems as they are inherently continuous \cite{pedersen2024neuromorphic}.
They have been demonstrated for speech detection \cite{martinelli2020spiking,du2024spiking,yang2024svad} and require gradient approximations due to the discontinuities of the spikes in backpropagation \cite{dampfhoffer2023backpropagation}.

%Neural Dynamics\cite{gerstner2014neuronal}

%Combining space-time representations\cite{laptev2005Space}

\section{Method}

\subsection{Wavelets}

Wavelets decompose a signal $f$ into $a,b$ components using the ``mother'' wavelet generating function 
\begin{equation}
    \psi(x; a, b) = |a|^{-1/2} \psi(\frac{x - b}{a})
\end{equation}
with a certain scale $a$ and temporal shift $b$
\begin{align}
    (T\, f)(x; a, b) &= \langle f, \psi(x; a, b) \rangle %\nonumber \\
                     = \int_{-\infty}^{\infty} \!\!\!f(x) |a|^{-1/2} \overline{\psi \left(\frac{x - b}{a} \right)} {\rm d} x
\end{align}
where $\psi$ satisfies $\int \psi(t) {\rm d}t = 0$ and $\langle \cdot, \cdot \rangle$ denotes the inner $L^2$ product \cite[p. 24]{daubechies2006Ten}, and $\overline{\ \psi \ }$ defines the complex conjugate of $\psi$.
By spanning the entire space of possible states, the resolution of identity ensures that the wavelet can represent any function transform \cite[(1.3.1)]{daubechies2006Ten}
\begin{equation} \label{eq:wavelet_recover}
    f(t) = C_\psi^{-1} \int_{-\infty}^\infty \int_{-\infty}^\infty {a^{-2}} 
           \langle f, \psi(t; a, b) \rangle\, \psi(t; a, b)\, {\rm d}a\, {\rm d}b,
\end{equation}
where $C_\psi$ is a constant dependent on $\psi$.

\subsection{Scale-space theory}

Scale-space theory parameterizes a signal $f\colon \mathbb{R} \to \mathbb{R}$ over its scale in a scale-space representation $L\colon \mathbb{R} \times \mathbb{R}_+ \to \mathbb{R}$ \cite{koenderink1984structure}.
This conveniently corresponds to the convolution of smoothing kernels $g(t;\ \tau)$, where $\tau$ is the scale parameter describing the variance of the smoothing kernels and $*$ denotes convolution \cite[(1.2, 1.3)]{lindeberg1994scale}
\begin{align}
    L(t;\ \tau) &= g(t;\ \tau) * f(x) %\nonumber \\
                = \int_{u = -\infty}^{\infty} g(u;\ \tau)\, f(t - u) {\rm d}u.
\end{align}
We require a causal relation between scales, such that variations are diminishing across scales, with features in smaller scales giving rise to features in larger scales, not vice versa.

\subsubsection{Scale space representation}
For any kernel $g$ to be variation diminishing, it must then adhere to the following Laplace transform
\cite[(3.60)]{lindeberg1994scale}
\begin{equation}\label{eq:limitkernel}
    \hat{\Psi}(s) = C\, e^{-\gamma s^2 + as} s^m \prod_{k=1}^{K = \infty} (1 + a_k s) e^{-a_k s}
\end{equation}
for $s \in \mathbb{C}$, where $C \in \mathbb{R}$ is a scaling constant, $\gamma \geqslant 0$ is an exponential scaling constant, $m \in \mathbb{R} \geqslant 0$ is a polynomial scaling factor, and $\sum_{k = 0}^{\infty} a_k < \infty$ is a convergent series.
Equation \eqref{eq:limitkernel} shows that in the limit of $K \to \infty$, we convolve $K$ kernels with varying delays $(1 + a_ks)$ and phase shifts $e^{-a_ks}$.
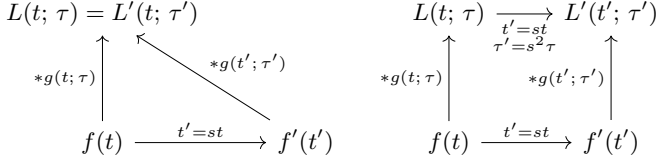
\begin{figure}
    \begin{subfigure}{0.2\textwidth}
    \begin{tikzcd}
      {L(t;\, \tau)} = {L'(t;\, \tau')} & \\
      \\
      f(t) & {f'(t')}
      \arrow["t'=st", from=3-1, to=3-2]
      \arrow["*g(t;\,\tau)", from=3-1, to=1-1]
      \arrow["*g(t';\, \tau')"', from=3-2, to=1-1]
      \end{tikzcd}
    \end{subfigure}
    \hfill
    \begin{subfigure}{0.2\textwidth}
        \begin{tikzcd}
        {L(t;\, \tau)} & {L'(t';\, \tau')} \\
        \\
        f(t) & {f'(t')} 
        \arrow["t'=st", from=3-1, to=3-2]
        \arrow["*g(t;\, \tau)", from=3-1, to=1-1]
        \arrow["*g(t';\, \tau')", from=3-2, to=1-2]
        \arrow["\substack{t'=st \\ \tau'=s^2\tau}"', from=1-1, to=1-2]
      \end{tikzcd}
    \end{subfigure}
  \caption{Transformational invariance (left) and covariance (right).
  When $L$ is invariant to a scaling in time by a factor $s$, the scaled representations $L$ and $L'$ do not distinguish between the original signal $f(t)$ and the transformed signal $f(t')$.
  The covariant representation retains the structure of the transformation, so that the right diagram commutes.
  %: the transformation applied to the signal, followed by mapping to $L'(t';\, \tau')$ is equivalent to first mapping the signal to $L(t;\, \tau)$ and then applying the corresponding transformation $s$ in the representation space.
  }
  \label{fig:covariance}
\end{figure}

In this work, we space the scaling factors $\tau_k$, in terms of the variance of the composed convolution kernel, logarithmically using powers of a distribution parameter $c > 1$ relative to a maximum scale level $\tau_{\text{max}}$ \cite[(18)]{lindeberg2016time}
\begin{equation} \label{eq:tau_k}
    \tau_k = c^{2(k - K)}\tau_{\text{max}}.
\end{equation}
We will refer to this transformed representation $\Psi$ \eqref{eq:limitkernel} as the ``time-causal limit kernel'' \cite[Sec. 3]{lindeberg2023time}.
%it must admit a bilateral Laplace transform \cite[(7)]{schoenberg1948variation}
%\begin{equation}
%    \int_{-\infty}^{\infty} e^{-sx} g(x) dx = \frac{\pm 1}{\Psi(s)},
%\end{equation}
%where $s$ is a scale factor.

% Two specific functions known to meet the above Laplace transform criterion are the Gaussian with a scaling factor $\tau$ \cite[(8)]{lindeberg2016time}
% \begin{equation} \label{eq:gaussian}
%    g_{\text{Gauss}}(x) = e^{-\tau x^2}
% \end{equation}
One function known to meet the above Laplace transform criterion is the truncated exponential with time constants $\mu_k$ \cite[(9)]{lindeberg2016time}
\begin{equation} \label{eq:truncated_exponential}
    g_{\text{exp}}(t, \mu_k) = \begin{cases}
        1/\mu_k\ \exp(-t/\mu_k)  & t > 0 \\
        0 & t \leqslant 0.
    \end{cases}
\end{equation}
The limit kernel corresponds to the composition of multiple such kernels in sequence, which yields a temporal mean $m_{\rm K} = \sum_k \mu_k$ and variance $\tau_{\rm K} = \sum_k \mu^2_k$ \cite[(10)]{lindeberg2023time}.
To ensure that the variance across these composed kernels amounts to $\tau$, the time constants $\mu$ distribute according to \cite[(20)]{lindeberg2016time}
\begin{align} \label{eq:mu1}
    \mu_1 &= c^{1 - K} \sqrt{\tau_{\rm max}} \\
    \mu_k &= c^{k - K - 1}\sqrt{c^2 - 1} \sqrt{\tau_{\rm max}} \qquad \text{for}\quad 2 \leqslant k \leqslant K. \label{eq:mu2}
\end{align}
Over the temporal domain, the time-causal limit kernel corresponds to the convolution of all possible combinations of the truncated exponentials for $k \to \infty$
\begin{equation}
    \Psi_{g_{\rm exp}}(t;\; \tau) = g_{\rm exp}(t;\, \mu_0) * g_{\rm exp}(t;\, \mu_1) * \dots * g_{\rm exp}(t;\, \mu_\infty).
\end{equation}

\subsubsection{Scale covariance}

The variation diminishing property ensures that scale-space representations preserve structure over scales. Using the time-causal limit kernel, the representation also obeys temporal scale covariance \cite[Secs. 3.1.2-3.1.3]{lindeberg2023time}.
Covariance,  also known as equivariance in some literature \cite{cohenGroupEquivariantConvolutional2016}, refers to the preservation of structure in some process, such as a neural network (see Figure \ref{fig:covariance}).
In the temporal case, consider the temporal scaling transformation $t_1 = s \, t_0$ and the corresponding scale parameter transformation $\tau_1 = s^2 \, \tau_0$ for scaling factor $s > 0$.
A scale-covariant representation satisfies
\begin{equation}
    L'(t_1;\, \tau_1) = L'(s \, t_0;\, s^2 \, \tau_0) = L(t_0;\, \tau_0),
\end{equation}
where the equality $L'(s \, t_0;\, s^2 \, \tau_0) = L(t_0;\, \tau_0)$ follows from the self-similarity property of the Gaussian kernel under appropriate scaling.
This means that $L'$ captures the same structural information as $L$, but the signal is observed at a different temporal scale.
Scale-covariant representations generalize scale-invariant ones: while invariant representations produce identical outputs regardless of input scale (Fig.\ \ref{fig:covariance}, left), covariant representations change predictably with the input transformation, preserving the relational structure between scales (Fig.\ \ref{fig:covariance}, right).

\subsubsection{Relation between scale-spaces and wavelets} \label{sec:scale-space-wavelets}

\begin{figure}
    \centering
    \includegraphics[width=\linewidth]{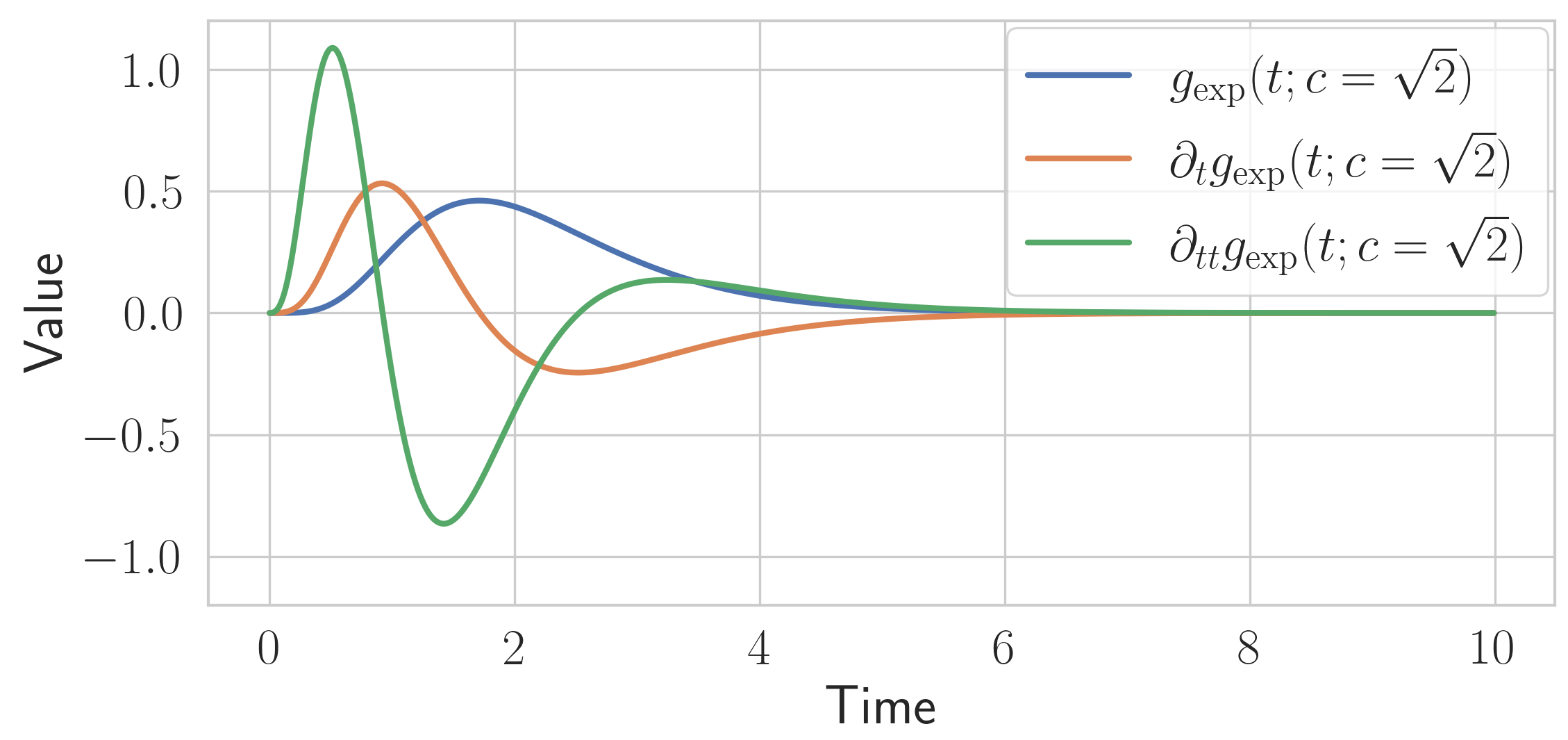}
 	\vspace{-0.4cm}
    \caption{A simulated approximation of the time-causal limit kernel \eqref{eq:limitkernel} approximated with the truncated exponential kernel $g_{\rm exp}$ \eqref{eq:truncated_exponential} and its first and second derivatives for $\tau_{\rm max} = 1$ for $K=7$.
    The kernels integrate to $(1 - 10^{-5})$, $2 \times 10^{-5}$, and $-2 \times 10^{-5}$, respectively.}
    \label{fig:limit_kernel}
\end{figure}

By parameterizing the limit-kernel $\Psi$ with the scale distribution parameter $c$, the scale-space representation $L$ for some signal $f$ that is causal in time at a point $t$ can be stated as \cite[(26)]{lindeberg2023time}
\begin{equation} \label{eq:scale-space}
    L(t; \tau, c) = \int_{u=0}^{\infty} \Psi(u; \tau, c) f(t - u) du.
\end{equation}
It is known that derivatives of the time-causal limit kernel can act as a mother wavelet over a continuous non-causal domain, provided they are normalized and satisfy $\int \psi(t) dt = 0$ \cite[(67)]{lindeberg2023time}
\begin{equation} \label{eq:differential_norm}
    W(t; \tau, c) = \frac{(\partial_{t^n}\Psi)(t; \tau, c)}{|| (\partial_{t^n} \Psi)(t; \tau, c)||},
\end{equation}
where $n \in \mathbb{N} \geqslant 1$ determines the $n$th derivative.
Figure \ref{fig:limit_kernel} visualizes the time-causal limit kernel with $K=7$ and is constructed following (\ref{eq:mu1}, \ref{eq:mu2}) with $c=\sqrt2$ and $\tau_{\rm max} = 1$ for $\mu$, i.e., $\{\mu_1, \mu_2, \mu_3, \dots, \mu_7\} = \{2^{-3}, 2^{-3}, 2^{-5/2}, \dots, 2^{-1/2}\}$, covering the interval $[0.125, 0.707]$.
For the numerical precision around $10^{-5}$,
the kernel $g_{\rm exp}$ integrates to 1, while the derivatives integrate to 0, demonstrating \eqref{eq:differential_norm}.

% Under a scaling-transformation of time $t' = c^jt$ for some integer $j$, this aligns with the scale-normalized temporal derivatives of $\Psi$
% %
% \begin{align}
%     \Psi'_n(t'; \tau' c') &= c^{jm(\gamma - 1)} \Psi_n(t; \tau, c) \\
%                           &= c^{j(1 - 1/p)} \Psi_n(t; \tau, c)
% \end{align}
% %
% where $m$ denotes the spatial derivative order, $\gamma$ is the scale-normalized power, and $p$ is the power in the $L_p$-norm \cite[eq. 71]{lindeberg2023time}.
% That is, by carefully choosing a discrete set of scales, scale-space representations fully capture the temporal scaling and translation operations of wavelets.

\begin{figure}[t]
    \centering
    \includegraphics[width=\linewidth]{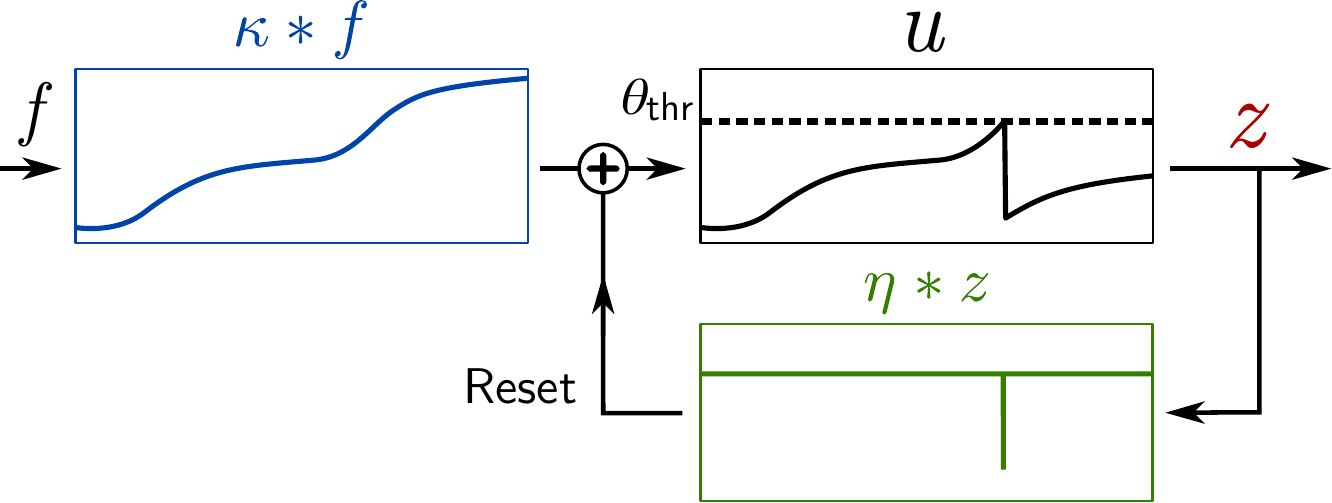}
	\vspace{-0.2cm}
    \caption{The spike response model (SRM) of a leaky integrate-and-fire system governed by $u$.
    SRM models neurons as compositions of linear systems, in this case an integrator $\kappa$, that generates spikes $z$ when $u \geqslant \theta_{\text{thr}}$, and a reset kernel $\eta$.}
    \label{fig:srm}
\end{figure}
\subsection{Spiking neural networks}

We leverage the relation between scale-space representations and spiking neural networks established in \cite{pedersen2025covariant} to construct a spiking neural network that can represent the wavelet system.
We choose the Spike Response Model (SRM) \cite[pp.\ 154]{gerstner2014neuronal} to describe generic neuron models with a single integral of composed linear kernels, see Fig.\ \ref{fig:srm}.
In the special case of the leaky integrate-and-fire model, we consider a leaky integrator term $\kappa$, a spike-train $z$, and reset kernel $\eta$ \cite[(6.28)]{gerstner2014neuronal}
\begin{equation} \label{eq:srm}
        u(t;\, \mu) = \int_0^\infty \big(\eta(s)\,z(t - s) + \kappa(s;\, \mu)\,f(t - s)\big)\,{\rm d}s.
\end{equation}
Consider a spike represented by a Dirac $\delta$ function, which satisfies $\int \delta(t)dt = 1$ and is zero for $t\neq0$.
The spiking output $z(t)$, is then the sum of $\delta$-functions emitted at times $t_{\text{f}}$  \cite[(1.14)]{gerstner2014neuronal}
\begin{equation}
    z(t) = \sum_{t_\text{f}}\delta(t - t_{\text{f}}).
\end{equation}
With the SRM kernels
\begin{equation}
    \mu \dot{\kappa}(t;\, \mu) = -\kappa(t;\, \mu) + \delta(t) \quad {\rm and} \quad \eta(t) = -\theta_{\rm thr}\, \delta(t), \label{eq:srm_eta}
\end{equation}
we get the explicit form
\begin{equation}
    u(t;\, \mu) = \int_0^\infty \frac{1}{\mu} e^{-s/\mu}\, f(t-s)\, {\rm d}s - \theta_{\rm thr}\, z(t).
\end{equation}
Via Laplace transform, this is equivalent to the differential equation
\begin{equation} \label{eq:lif_ode}
    \mu \dot{u} = -u(t) + f(t) - \theta_{\rm thr}\, z(t),
\end{equation}
which is the standard LIF equation with instantaneous reset \cite[Sec. 1.3]{gerstner2014neuronal}.
% That is, membrane voltage $u(t)$ is the sum of the input $f(t)$ convolved with a leaky integrator $\kappa(t)$, for which we use $g_{\rm exp}(t)$ \eqref{eq:truncated_exponential}, and a reset term that sets the membrane voltage to 0 when $u$ crosses the threshold $\theta_{\rm thr}$
% %
% \begin{equation} \label{eq:li_dot}
%         \mu\, \dot{u} = -u + f(t).
% \end{equation}
% The reset kernel $\eta$ resets the membrane potential to 0 when $u$ crosses the threshold $\theta_{\rm thr}$
% %
% \begin{equation} \label{eq:reset_instant}
%         % u(t) = \begin{cases}
%         %         0 & z(t) = 1 \\
%         %         u(t)           & z(t) = 0.
%         % \end{cases}
% \mu\, \dot{u} = -u + f(t) - u(t)z(t).
% \end{equation}
Solving equation \eqref{eq:lif_ode} via the SRM yields a scale-space representation at scale $\mu$ for some input signal $f$ \cite[(23)]{pedersen2025covariant}
\begin{equation} \label{eq:lif_srm}
        L(t; \mu) = -\theta_{\text{thr}}\ e^{-(t-t_{\rm f})/\mu} + \int_{z=0}^\infty f(t - z)\, \frac{1}{\mu}e^{-z/\mu}\, {\rm d}z,
\end{equation}
where $t_{\rm f}$ is the time of the previous spike.
Considering temporal scaling operations $t' = st$ and $t'_f = st'_f$ for the functions $f'(t') = f(t)$, where $u' = su$, ${\rm d}u' = s{\rm d}u$, and $\mu' = s\mu$.
The scaling factors cancel, demonstrating scale covariance for the LIF neuron model \cite[(24)]{pedersen2025covariant}
\begin{align} \label{eq:srm_temporal_covariance}
     & L(t'; \mu')
    = -\theta_{thr}\ e^{-(t'-t_{\rm f}') / \mu'} + \int_{z'=0}^\infty  \!\!\!\!\!\! f'(t' - z')\, \frac{1}{\mu'}e^{-z' / \mu'} {\rm d}z'                       \nonumber \\
  &= -\theta_{thr}e^{-(t-t_{\rm f}) / \mu} \! + \!\!\int_{z=0}^\infty \!\!\!f(t - z)\, \frac{1}{\mu}e^{-z/\mu}\, {\rm d}z
  = L(t; \mu).
\end{align}

\begin{figure}[t]
    \centering
    \includegraphics[width=\linewidth]{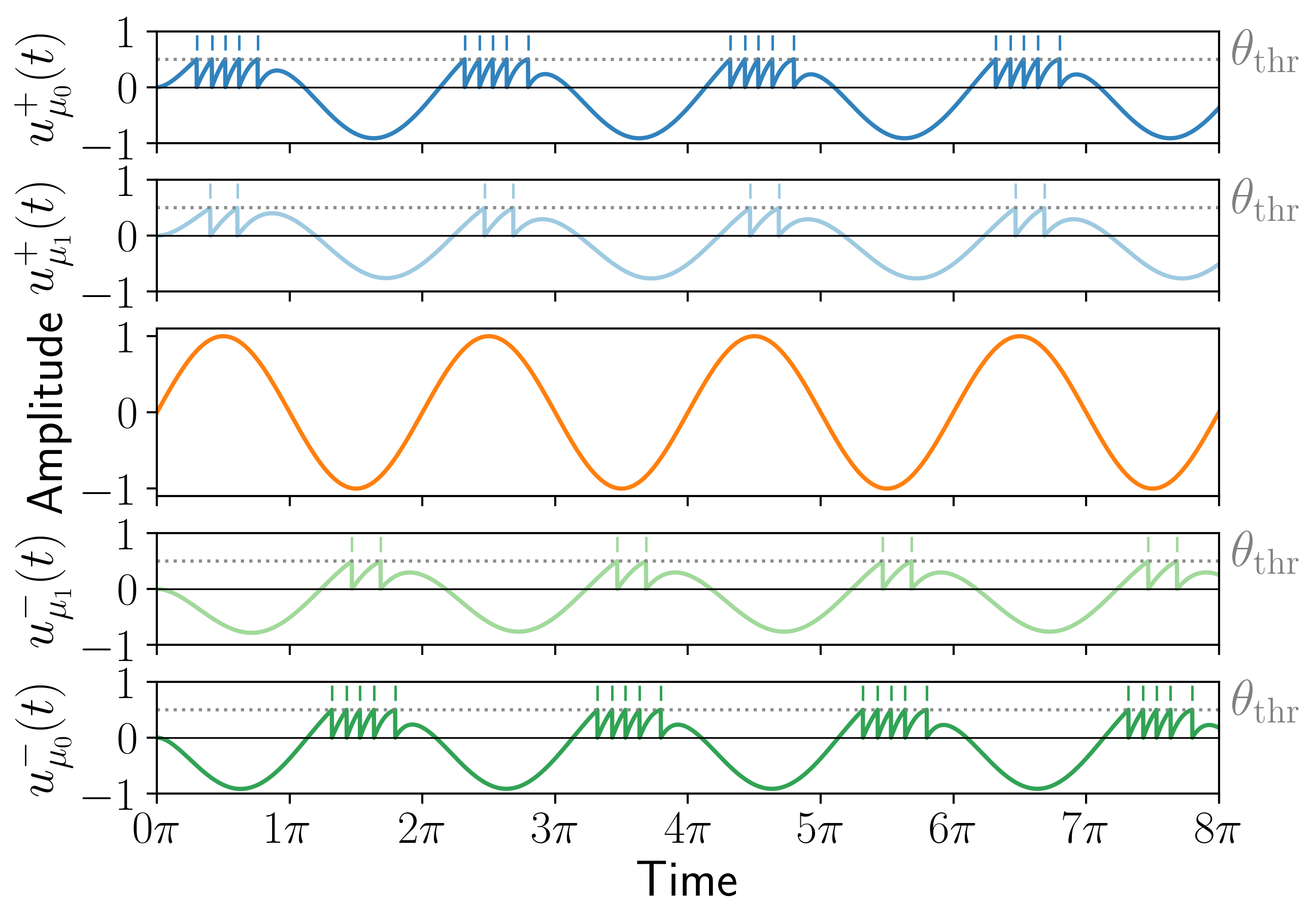}
	\vspace{-0.8cm}
    \caption{Two-channel spiking kernel representation for a signal across temporal scales ($\mu_0$ and $\mu_1$).
    Signal $f(t) = \sin(t)$ (orange) is integrated \eqref{eq:srm} by positive neurons (blue) and negative neurons (green), with $\theta_{\rm thr} = 0.5$.
    }
    \label{fig:two-channel}
\end{figure}

\begin{figure*}
    \centering
    \includegraphics[width=\textwidth]{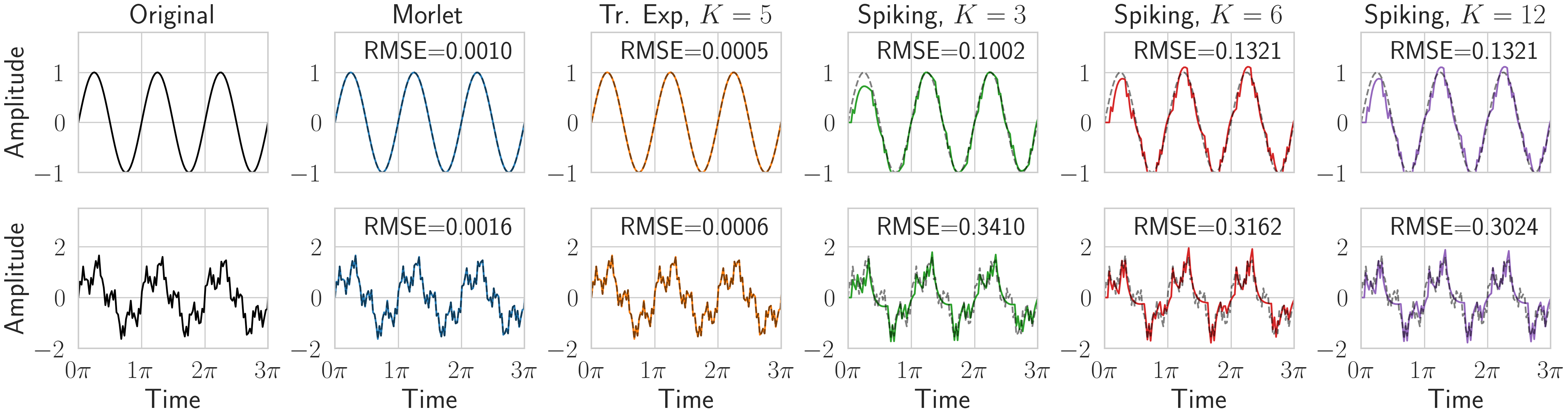}
	\vspace{-0.6cm}
    \caption{Reconstruction of a sinusoidal (top) and a composite sinusoidal (bottom) with different wavelets.
    The left panels show the raw signals. 
    From left to right, the wavelets are: Morlet, truncated exponential \eqref{eq:truncated_exponential} with $K = 5$, and spiking wavelets from \eqref{eq:positive_negative_decomposition} with $K = \{3, 6, 12\}$.
    The original signal is superimposed on the reconstructions as a dashed line for comparison.
    }
    \label{fig:reconstruction}
\end{figure*}

\subsection{Spiking mother wavelet}

Wavelet representations require that signs are preserved (transformation of $f$ should be $T(f)$, transformations of $-f$ should be $-T(f)$).
However, outgoing spikes do not carry sign information, they inherently discard negative values.
Similar to rectifying circuits and Dale's principle \cite{eccles1954cholinergic}, where neurons are either exclusively excitatory or exclusively inhibitory, we address the sign problem by constructing a two-channel representation at some scale $\mu$ with two integrative terms
\begin{align} \label{eq:spiking_channels}
    \mu\,\dot{u}^+ &= -u^+ + f(t) \nonumber \\
    \mu\,\dot{u}^- &= -u^- -f(t).
\end{align}
When $u^{\pm}$ crosses $\theta_{\rm thr}$, the neuron emits a spike \eqref{eq:srm} and resets \eqref{eq:srm_eta}.
This provides us with ``positive'' and ``negative'' channels that produce spike trains $z^+(t)$ and $z^-(t)$.

To construct a multi-scale representation, we implement the kernel $\kappa$ at each scale $k$ as a difference of exponentials
\begin{equation} \label{eq:kappa_difference}
    \kappa(t;\, \mu_k) = \frac{1}{C_\kappa}\left(g_{\rm exp}(t;\, \mu_{k+1}) - g_{\rm exp}(t;\, \mu_k)\right),
\end{equation}
where $C_\kappa = (\mu_{k+1} - \mu_k)/(\mu_k \mu_{k+1})$ normalizes the kernel, and $\mu_k < \mu_{k+1}$ are adjacent time constants in the scale hierarchy.
This corresponds to two coupled LIF neurons \eqref{eq:lif_ode} with time constants $\mu_k$ and $\mu_{k+1}$, forming the membrane potential $u(t) = (u_{\rm slow} - u_{\rm fast})/C_\kappa$ and creating bandpass filters that satisfy the zero-mean requirement for wavelet admissibility.

At each scale $\mu_k$, we reconstruct the signal by convolving the spike trains with the difference kernel
\begin{equation}
    M^{\pm}(t;\, \mu_k) = \int_0^{\infty} \kappa(s;\, \mu_k)\,  z^{\pm}(t-s)\, {\rm d}s.
\end{equation}
The full signal representation at scale $\mu_k$ combines both polarity channels:
\begin{equation} \label{eq:positive_negative_decomposition}
    M(t;\, \mu_k) = M^+(t;\, \mu_k) - M^-(t;\, \mu_k).
\end{equation}
The difference kernel exhibits approximate scale covariance when both time constants scale proportionally, ($\mu'_k = s\mu_k$, $\mu'_{k+1} = s\mu_{k+1}$), preserving the bandpass structure across scales.
Since the sum of covariant functions is covariant, $M(t;\, \mu_k)$ retains the scale-covariant properties required to act as a wavelet representation.
Since $M$ exclusively relies on truncated exponentials, this is entirely representable with leaky integrators as physical primitives.
Consequently, the output of the spiking polarity channels becomes a discretized, but scale-covariant, version of the underlying wavelets.

Figure \ref{fig:two-channel} visualizes spiking LIF channels at varying temporal scales using the Python library Norse \cite{norse2021}.
The sinusoidal signal (black) is integrated into spiking neurons for 2 positive channels and 2 negative channels, all with the spiking threshold $\theta_{\rm thr} = 0.5$. 
%The sinusoidal signal (black) is integrated into 2 spiking neurons with a positive spiking threshold and 2 neurons with a negative spiking threshold.
The threshold determines the point at which the integrated signal results in a spike and an immediate reset.
The speed at which they integrate and fire is determined by the time parameter $\{\mu_0, \mu_1\}$, where $\mu_1$ yields the slowest integration.
We visualize the integration of the slowest timescale $\mu_1$ for positive (blue) and negative (green) channels, yielding two spikes at every signal peak.
Notice the reset following each spike.

\section{Experiment}

We demonstrate the spiking neural network wavelets on a sinusoidal  $f(t)=\sin(t)$ and an arbitrary composite sinusoidal $f(t)=\sin(2\pi \cdot 0.5\, t) + 0.5\sin(2\pi\, 2\, t) + 0.3\sin(2\pi \, 8\, t)$ over time $t$, see Fig.\ \ref{fig:reconstruction}.
As a reference, we used the Morlet wavelet \cite[(3.3.26)]{daubechies2006Ten}
\begin{equation}
    \psi_{\rm morlet}(t) = \frac{1}{\sqrt{\pi\sigma^2}} \exp\left(-\frac{t^2}{2\sigma^2}\right) \exp(i\omega_0 t).
\end{equation}
Furthermore, we compare the time-causal limit kernel \eqref{eq:limitkernel} ($K=3$), and spiking wavelets ($K \in \{3, 6, 12\}$).
For spiking wavelets \eqref{eq:positive_negative_decomposition}, coefficients are computed at each scale by counting spikes in temporal bins of width $\Delta t$.
The coefficient at time $t$ and scale $\mu_k$ is
\begin{equation}
    W_{\rm spike}(t;\, \mu_k) = \left(n_{\rm pos}(t) - n_{\rm neg}(t)\right)/\sqrt{\mu_k},
\end{equation}
where $n_{\rm pos}(t)$ and $n_{\rm neg}(t)$ are the spike counts in the positive and negative channels within the bin centered at $t$, and the normalization factor $1/\sqrt{\mu_k}$ accounts for scale-dependent energy.
The time constants for the filters are initialized using (\ref{eq:mu1}, \ref{eq:mu2}) where $c = \sqrt{2}$ and $\tau_{\rm max} = 3.4$.
As $K$ increases, we reduced $c$ to $3$ for $K=6$ and $1.6$ for $K=12$ to improve reconstruction accuracy.
The threshold for the spiking models is $\theta_{\rm thr} = 0.1$.

The truncated exponential wavelet achieves best reconstruction, followed by Morlet, and spiking wavelets.
The spiking wavelet performance plateaus beyond $K=3$ channels, suggesting that the spike-count encoding scheme via \eqref{eq:spiking_channels} captures the frequency components without requiring additional channels. 
The plateau hints at the need for more sophisticated spike-based and spike-time encoding strategies that better exploit additional channels, including cross-channel dependencies, for richer signal representations.
Our approach demonstrates that neural integration can transform signals into wavelet representations, providing a foundation for investigating how biological constraints and neural dynamics influence wavelet analysis.

\section{Discussion}
We established that spiking neural networks form a mother wavelet in the theoretical limit.
This provides a mathematically grounded connection between classical signal processing and neuromorphic hardware, with dual benefits: information efficiency through temporally sparse spikes capturing multi-scale structure, and energy efficiency through neuromorphic substrates that consume energy only during spike events.
The scale-covariant properties ensure these advantages hold across temporal scales.
For truncated exponential and spiking wavelets, simple signals are reasonably reconstructed, with reconstruction quality improving with the number of scales.
Our findings are based on the spike-response model, which is sufficiently expressive to capture many other neuron models that might provide other, and better, bases.
Threshold adaptation, where the threshold increases under repeated firing, would be interesting to study, since it could yield increased sparsity without loss.

%\newpage
\bibliographystyle{IEEEbib}
\bibliography{refs}
\end{document}